\documentclass[letterpaper, 10 pt, conference]{ieeeconf}  

\IEEEoverridecommandlockouts                              

\usepackage{graphics} 
\usepackage{epsfig} 
\usepackage{times} 
\usepackage{amsmath} 
\usepackage{amssymb}  
\usepackage{resizegather}
\usepackage{mathtools}
\usepackage{gensymb}
\usepackage{subcaption}
\usepackage{tabularx,booktabs}
\usepackage{changes}
\newcolumntype{Y}{>{\centering\arraybackslash}X}

\title{\LARGE \bf{Multi-IMU Proprioceptive State Estimator for Humanoid Robots}}

\author{\resizebox{\linewidth}{!}{Fabio Elnecave Xavier$^{1,2}$, Guillaume Burger$^1$, Marine Pétriaux$^1$, Jean-Emmanuel Deschaud$^2$ and François Goulette$^{2,3}$}
\thanks{$^1$Wandercraft, 75004 Paris, France}%
\thanks{$^2$Centre for Robotics, Mines Paris, PSL University, 75006 Paris, France}%
\thanks{$^3$U2IS, ENSTA Paris, Institut Polytechnique de Paris, 91120 Palaiseau, France}%
\thanks{\tt\small fabio.elnecave\_xavier@minesparis.psl.eu}%
}

\newcommand{\R}{\mathbb{R}}
\newcommand{\Stwo}{{\mathbb{S}^2}}
\newcommand{\manM}{\mathcal{M}}
\newcommand{\Exp}{\text{Exp}}
\newcommand{\Log}{\text{Log}}
\newcommand{\Rot}{\text{Rot}}
\newcommand{\ehat}[1]{#1^\wedge}
\newcommand{\evee}[1]{#1^\vee}
\newcommand{\argdot}{\makebox[1ex]{\textbf{$\cdot$}}}

\begin{document}
\maketitle
\thispagestyle{empty}
\pagestyle{empty}

\begin{abstract}

Algorithms for state estimation of humanoid robots usually assume that the feet remain flat and in a constant position while in contact with the ground. However, this hypothesis is easily violated while walking, especially for human-like gaits with heel-toe motion. This reduces the time during which the contact assumption can be used, or requires higher variances to account for errors. In this paper, we present a novel state estimator based on the extended Kalman filter that can properly handle any contact configuration. We consider multiple inertial measurement units (IMUs) distributed throughout the robot's structure, including on both feet, which are used to track multiple bodies of the robot. This multi-IMU instrumentation setup also has the advantage of allowing the deformations in the robot’s structure to be estimated, improving the kinematic model used in the filter. The proposed approach is validated experimentally on the exoskeleton Atalante and is shown to present low drift, performing better than similar single-IMU filters. The obtained trajectory estimates are accurate enough to construct elevation maps that have little distortion with respect to the ground truth.

\end{abstract}

\section{INTRODUCTION}

Humanoid robots require accurate estimates of their floating base pose and velocity for control, motion planning, and mapping tasks. Because these variables cannot be measured directly, their estimates must be computed via the fusion of data from other sensors. Even though vision-based methods have been proposed to solve this problem, solutions that only use proprioceptive sensors have the advantage of allowing the estimator to run at a high frequency while being unaffected by changes in lighting or environmental conditions. A common approach in this case is to combine the measurements of an inertial measurement unit (IMU), leg kinematics, and contact information using an extended Kalman filter (EKF).

This idea was first explored by \cite{bloesch2012state}, which formulated a state estimator for a legged robot with an arbitrary number of point feet. The main idea of their method is to include the position of the footholds in the state of the filter. In the prediction step, the pose and velocity of the floating base are computed by integrating the measurements from the IMU's accelerometer and gyroscope, and the feet in contact with the ground are assumed to remain static. Then, in the correction step, the state estimate is updated using as a measurement the position of the feet with respect to the base, which is computed from the joint encoder readings using the robot's kinematic model. In \cite{rotella2014state}, this method was extended to humanoid robots (which usually have large, flat feet for stability) by adding the orientation of the feet to the state. In this case, the measurement used in the correction step is the full pose of the feet with respect to the base.

The two filters described use a framework known as error state extended Kalman filter (ESEKF) \cite{sola2017quaternion} to represent the uncertainty and update the estimates of variables representing three-dimensional (3D) orientations. In \cite{hartley2020contact}, a variation of \cite{bloesch2012state} based on the invariant extended Kalman filter (InvEKF) \cite{barrau2016invariant} is proposed. In the InvEKF, if the full state can be defined on a matrix Lie group and its dynamics satisfy a group-affine property, it is possible to obtain an error propagation that does not depend on the current state estimate. Consequently, the filter presents better convergence and consistency properties. Two different variations of this formulation for the case of humanoid robots with flat feet were proposed by \cite{qin2020novel} and \cite{ramadoss2021diligent}, which differ in how the orientation of the feet is represented in the state.

\begin{figure}
    \raggedright
    \begin{subfigure}[c]{0.40\linewidth}
        \centering
        \includegraphics[width=\linewidth]{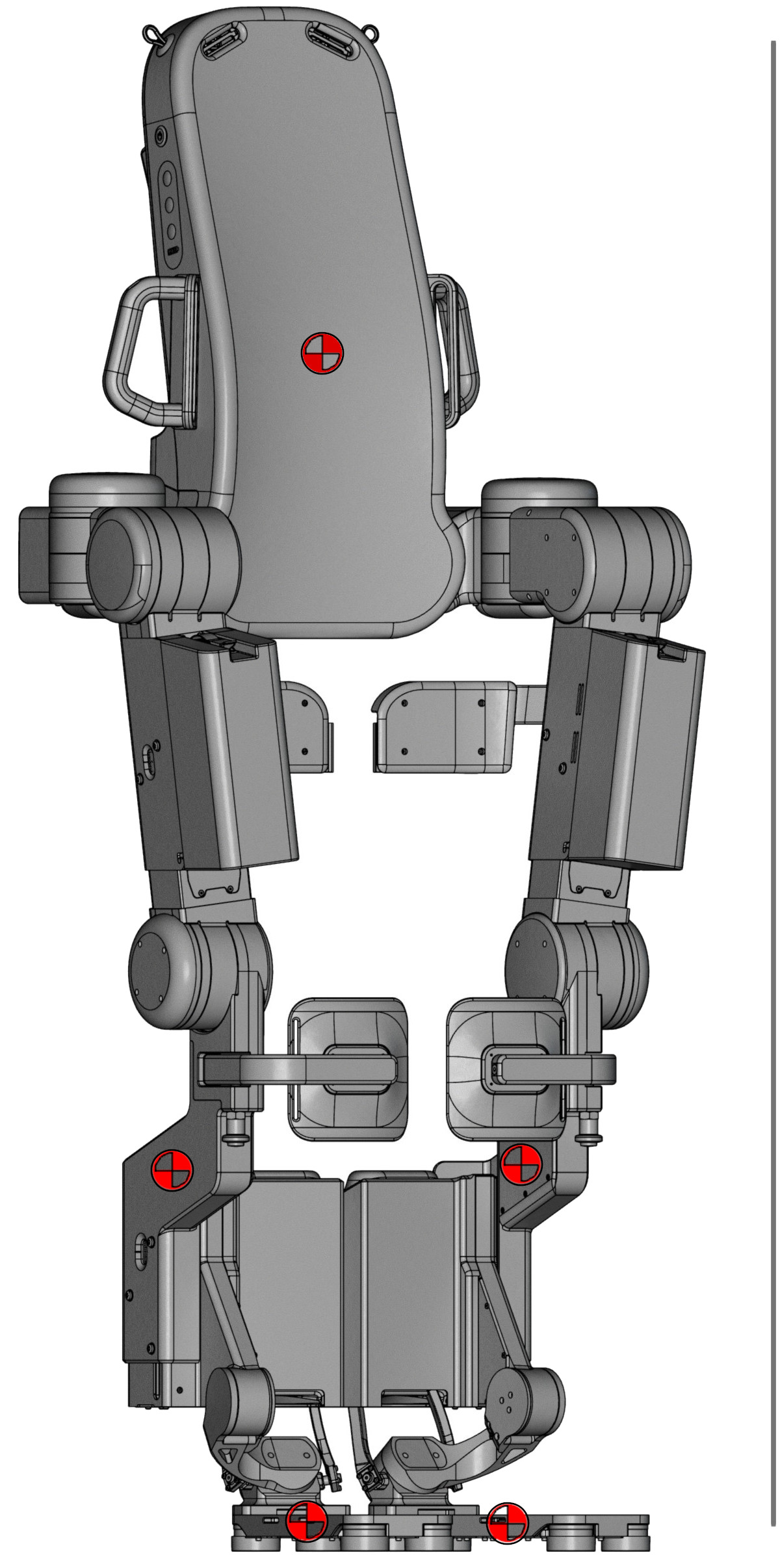}
    \end{subfigure}
    \hfill
    \begin{subfigure}[c]{0.58\linewidth}
        \centering
        \includegraphics[width=\linewidth]{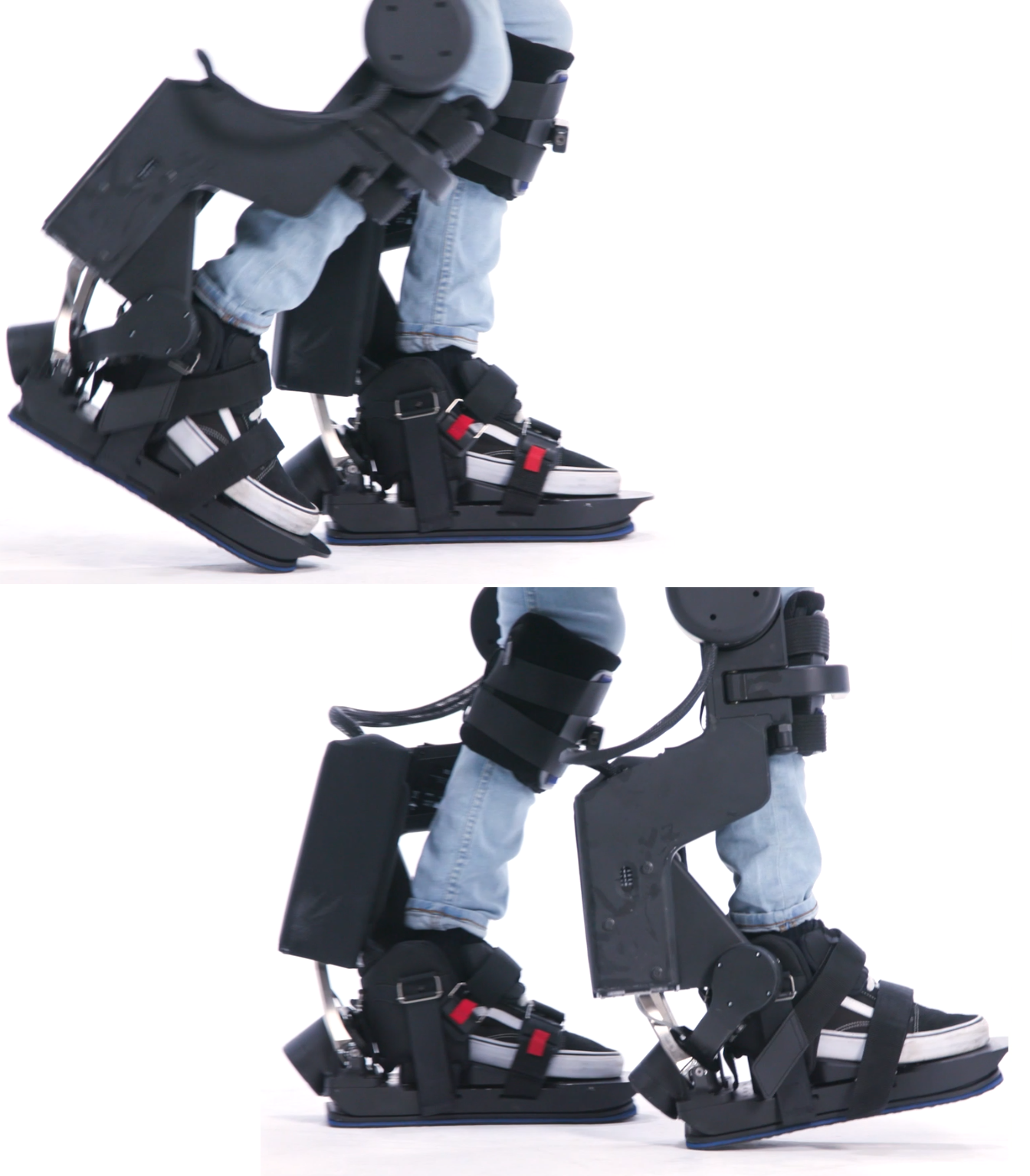}
    \end{subfigure}
    \setlength{\belowcaptionskip}{-15pt}
    \caption{On the left, CAD model of the exoskeleton Atalante showing in red the location of its five IMUs. On the right, illustration of a gait pattern with heel-toe motion. The multi-IMU state estimator is able to exploit contact information even when the foot is not flat on the floor.}
    \setlength{\belowcaptionskip}{0pt}
    \label{fig:atalante_heel_toe}
\end{figure}

All state estimators for humanoid robots presented above consider that the only link instrumented with an IMU is the floating base. Therefore, since they lack the necessary measurements for modelling it otherwise, they must assume that the pose of the feet is constant while in contact with the ground. The accuracy of the estimator depends on the capacity of reliably detecting when such stable contact occurs, which is not a trivial task depending on the available sensors. 

In this paper, we introduce a novel EKF state estimator that does not rely on this assumption. For this, we require the robot to be instrumented with multiple IMUs, including, necessarily, one on each foot. By modelling the movement of the feet, we are able to handle any contact configuration, and can therefore exploit the contact information over a longer time interval. This is especially significant for anthropomorphic gaits that include heel-toe motion, that is, in which the heel both touches and leaves the floor before the toes (see Fig. \ref{fig:atalante_heel_toe}). The multi-IMU setup is also used to estimate the deformations in the robot's structure, as in \cite{vigne2022movie}, improving the accuracy of the kinematic model used in the filter. The proposed estimator is derived using the ESEKF, and it can be seen as an extension of \cite{rotella2014state} to the case of multiple IMUs.

We demonstrate the effectiveness of our method in real-world experiments with the medical leg exoskeleton Atalante, and we show that it always outperforms a single-IMU estimator. The multi-IMU filter provides more accurate trajectory estimates for the two gait patterns analysed. The improvement is especially significant in the vertical direction, where the proposed estimator presents very low drift, even in the presence of heel-toe motion. By attaching a solid-state LiDAR to the exoskeleton, we also show that the trajectories estimated using our method are accurate enough to construct elevation maps that closely resemble the terrain ground truth.

The structure of the paper is as follows. In Section \ref{sec:overview}, we provide a short overview of our method. The ESEKF framework and the notation we will employ are carefully defined in Section \ref{sec:esekf}. Then, in Section \ref{sec:algorithm}, we present the state estimator algorithm in detail. Finally, the experimental results are discussed in Section \ref{sec:results}.

\section{OVERVIEW OF THE ESTIMATOR}
\label{sec:overview}

The objective of the multi-IMU state estimator is to compute the position, orientation, and velocity of multiple links of a humanoid robot --- each of them instrumented with an IMU --- in a fixed inertial reference frame. We assume that the robot is also equipped with joint encoders and that it can compute the location of the center of pressure of the feet. We follow the same basic structure as other EKF-based estimators: in the prediction step, the measurements from the IMUs are integrated independently to provide a first estimate of the pose and velocity of each link; then, in the correction step, these estimates are updated using the robot's kinematic model.

We employ two different prediction models, which are chosen for each link depending on their contact status. For the \textit{floating links}, that is, all links not in contact with the ground at the moment, we adopt the standard inertial navigation model used by \cite{rotella2014state}. On the other hand, the \textit{contact links} --- meaning the foot or feet in contact with the ground --- are modelled as instantly rotating around their center of pressure. This is a reasonable contact hypothesis, even in the presence of strong heel-toe motion or rotational slippage. In this case, the measurement of the gyroscope is sufficient to compute the velocity of the link, and the accelerometer will only be used indirectly in the correction step.

For the correction step of the filter, we construct an extended kinematic model of the robot that considers the deformations in the mechanical structure, which are estimated by the observer MOVIE \cite{vigne2022movie}. MOVIE fuses the measurements from the robot's IMUs and encoders to produce an estimate of the \textit{tilt} of each IMU, that is, the unit vector that points in the direction of gravity, expressed in the sensor frame. Thus, the tilt corresponds to the observable component of the IMU's orientation. Then, starting with one stance foot and traversing the kinematic chain sequentially, the tilt of each IMU estimated by MOVIE is compared to its orientation according to the robot's standard kinematic model, which assumes the structure is perfectly rigid. The difference between these values is interpreted as a mechanical deformation, which we model as a rotation around a known point in the structure. 

Based on MOVIE and on this extended kinematic model, we define the two different measurements used in the correction step. The first is the relative pose of each floating link with respect to each contact link. This can be seen as a natural generalization to the multiple-IMU case of the measurements used in \cite{rotella2014state}. It ensures the estimated poses of the multiple links remain spatially consistent, even though they are updated independently in the prediction step. However, it does not capture any information about the absolute orientation of the links. For this, we also include the tilt of each contact link as a measurement. This allows us to consider the readings of their accelerometers, which were ignored during the prediction step.

\section{ESTIMATION FRAMEWORK}
\label{sec:esekf}

The ESEKF has become a standard method for estimating 3D orientations in robotics (\cite{vitali2020robust,markovic2022error}). However, its application is not restricted to SO(3). In our case, it will be used to manipulate both 3D orientations and IMU tilts, which are defined on the manifold $ \Stwo $, i.e., the set of unit vectors in $ \R^3 $. In this section, we provide a self-contained exposition of the ESEKF at a level of abstraction that suits our needs. Presenting it in this general setting and defining the necessary notation will clarify the presentation of the state estimation algorithm.

\subsection{Mathematical preliminaries}

Let $ \manM $ be an $n$-dimensional manifold. We define a perturbation operator $ \oplus_\manM: \manM \times \R^n \to \manM $ such that, for every $ x \in \manM $, $ x \oplus_\manM 0 = x $, and $ \delta \mapsto x \oplus_\manM \delta $ is a parametrization of a neighborhood of $ x $ for a sufficiently small $ \delta \in \R^n $. This allows us to work with local parametrizations of the manifold around each point instead of a global minimal parametrization, which could contain singularities (e.g. Euler angles for SO(3), polar and azimuthal angles for $ \Stwo $). We also define the inverse operator $ \ominus_\manM: \manM \times \manM \to \R^n $ so that $ x \oplus_\manM (y \ominus_\manM x) = y $ for sufficiently close $ x, y \in \manM $. This allows us to transcribe many algorithms from $ \R^n $ to $ \manM $ simply by replacing the vector space operators $ + $ and $ - $ with $ \oplus_\manM $ and $ \ominus_\manM $ (see \cite{hertzberg2008framework} for more details). After this section, we will omit the subscript and denote the operators by $ \oplus $ and $ \ominus $ to simplify the notation.

When the manifold is also a Lie group, these operators can be easily defined using the group operation \cite{sola2018micro}. For example, we will use the following definitions for SO(3): 
\begin{align}
    R \oplus_{\text{SO(3)}} \delta &= R~\Exp(\delta) = R \exp(\ehat{\delta}) \\
    R_1 \ominus_{\text{SO(3)}} R_2 &= \Log({R_2}^T R_1) = \evee{\log({R_2}^T R_1)}
\end{align}
where $ \exp $ and $ \log $ are the standard matrix exponential and logarithm, $ \ehat{(\argdot)} $ is the operator that maps a vector of $ \R^3 $ to a skew-symmetric matrix such that $ \ehat{a} b = a \times b $, and $ \evee{(\argdot)} $ is its inverse.

This idea cannot be extended to $ \Stwo $, as it does not possess a group structure. In this case, inspired by \cite{he2021kalman}, we will use the operators
\begin{align}
    u \oplus_\Stwo \delta &= \Exp(B(u) \delta) u \label{eq:oplus_s2} \\
    v \ominus_\Stwo u &= B(u)^T \Log(\Rot(u, v))
\end{align}
where $ B(u) $ is a $ 3 \times 2 $ matrix with orthonormal columns that span the two-dimensional subspace orthogonal to $ u $, and $ \Rot(u, v) $ is the rotation matrix that corresponds to the smallest rotation that transforms $ u $ into $ v $.\footnote{This is well defined for all $ u, v \in \Stwo $ that are not diametrically opposite each other. Using Rodrigues' rotation formula, we obtain that $ \Rot(u,v) = I + \ehat{(u \times v)} + \frac{1}{1 + u \cdot v} [\ehat{(u \times v)}]^2 $.}

The operator $ \oplus $ provides a natural way to define a normal distribution for a random variable $ X $ in a manifold. The idea is to the define the distribution on $ \R^n $ and then map it to the manifold using the perturbation operator. We will say that $ X $ is normally distributed with mean $ \hat{x} \in \manM $ and covariance matrix $ P $ if we can write
\begin{equation}
    X = \hat{x} \oplus_\manM \delta X
\end{equation}
where $ \delta X $ is a normally distributed perturbation in $ \R^n $ with zero mean and covariance matrix $ P $.

These operators also allow us to compute the equivalent of a linear approximation of a function $ \phi : \manM_1 \to \manM_2 $, where $ \manM_1 $ and $ \manM_2 $ are, respectively, an $n$-dimensional and an $m$-dimensional manifold. For $ x \in \manM_1 $ and a small $ \delta \in \R^n $, we have
\begin{equation}
    \phi(x \oplus_{\manM_1} \delta) \approx \phi(x) \oplus_{\manM_2} (J(x) \delta)
\end{equation}
where the $ m \times n $ matrix $ J(x) $ corresponds to the Jacobian of $ \delta \mapsto \phi(x \oplus_{\manM_1} \delta) \ominus_{\manM_2} \phi(x) $ at $ \delta = 0 $. We will refer to $ J(x) $ as the Jacobian of $ \phi $ evaluated at $ x $.

\subsection{The Error State Extended Kalman Filter}

We consider the problem of estimating the state $ x $ of a system defined on an $n$-dimensional manifold $ \manM_s $, having access only to a measurement $ y $ belonging to an $m$-dimensional manifold $ \manM_m $. The system is described by the equations
\begin{align}
    \dot{x} &= f(x, \eta) \label{eq:system_equation} \\
    y &= g(x, \nu) \label{eq:measurement_equation}
\end{align}
where $ \eta $ and $ \nu $ are zero-mean Gaussian white noise signals in $ \R^p $ and $ \R^q $, with covariance matrices $ H $ and $ R $, respectively. The measurements and the estimation are made with a sampling time $ \Delta t$.

At each discrete time instant $ k $, we model our estimate of the state as a normal variable
\begin{equation}
    X_k = \hat{x}_k  \oplus \delta X_k \label{eq:normal_distribution}
\end{equation}
with covariance matrix $ P_k $. The perturbation $ \delta X_k $ corresponds to an estimate of the \textit{error state}, that is, the difference between the actual value of the state and our estimate. Because it is a random variable in $ \R^n $, we will employ the usual prediction and correction equations of the EKF to update it and then map this update back to the manifold using the $ \oplus $ operator.

\subsubsection{Prediction step}

Let our estimate of the state at instant $ k-1 $ be $ X_{k-1} = \hat{x}_{k-1} \oplus \delta X_{k-1} $ with covariance matrix $ P_{k-1} $. In the prediction step of the ESEKF, we will obtain a first estimate of $ x_k $ using \eqref{eq:system_equation} only, which we will denote $ X_k^- = \hat{x}_k^- \oplus \delta X_k^- $, with covariance $ P_k^- $. For such, we write the state as
\begin{equation}
    x = \hat{x} \oplus \delta x,
    \label{eq:state_decomposition}
\end{equation}
and we derive the differential equation that describes how each of these components evolves in time. We would like $ \hat{x} $ to follow the expected evolution of the state while the error state $ \delta x $ captures the uncertainties due to noise and to the previous estimates. Thus, it is natural to consider that $ \hat{x} $ follows a noiseless version of \eqref{eq:system_equation}
\begin{equation}
    \dot{\hat{x}} = f(\hat{x}, 0).
    \label{eq:noiseless_system_equation}
\end{equation}
This allows us to compute $ \hat{x}_k^- $ by integrating this differential equation with the initial condition $ \hat{x}_{k-1} $ for a time interval $ \Delta t $.

The differential equation for $ \delta x $ is obtained by differentiating both sides of \eqref{eq:state_decomposition} with respect to time and substituting the expressions \eqref{eq:system_equation} and \eqref{eq:noiseless_system_equation}. Because differentiation does not behave linearly with respect to $ \oplus $, a general expression for this differential equation would be overly complicated; it is easier to compute it from the definition for each system. Then, by linearizing it with respect to $ \delta x $ and $ \eta $ around the origin, we obtain an equation in the form
\begin{equation}
    \dot{\delta x} = F(\hat{x}) \delta x + G(\hat{x}) \eta, \label{eq:error_state_ode}
\end{equation}
where $ F(\hat{x}) $ and $ G(\hat{x}) $ are $ n \times n $ and $ n \times p $ Jacobian matrices, respectively.

Finally, by making the approximations $ F(\hat{x}) \approx F(\hat{x}_k^-) = F_k $ and $ G(\hat{x}) \approx G(\hat{x}_k^-) = G_k $ on the interval between instants $ k - 1 $ and $ k $, we obtain a closed-form discretization of \eqref{eq:error_state_ode} as
\begin{equation}
    \delta x_k = A_k \delta x_{k-1} + w_k, \label{eq:error_state_discrete}
\end{equation}
where $ A_k = \exp{(F_k \Delta t)} $ and $ w_k $ is a Gaussian vector with zero mean and covariance matrix
\begin{equation}
\resizebox{0.89\linewidth}{!}{$%
    Q_k = \int_0^{\Delta t} \exp{(F_k(\Delta t - t))} G_k H G_k^T \exp{({F_k}^T(\Delta t - t))} dt.  
$}
\end{equation}
In practice, we prefer using the first-order approximations
\begin{align}
    A_k &\approx I + F_k \Delta t \label{eq:Ak_approx} \\
    Q_k &\approx G_k H {G_k}^T \Delta t  \label{eq:Qk_approx}
\end{align}
which are valid for a small $ \Delta t $. The linear equation \eqref{eq:error_state_discrete} allows us to compute an approximation for the distribution of $ \delta X_k^- $. Indeed, if $ \delta X_k^- = A_k \delta X_{k-1} + w_k $, it follows that it is normally distributed with zero mean and covariance matrix
\begin{equation}
    P_k^- = A_k P_{k-1} {A_k}^T + Q_k. \label{eq:covariance_prediction}
\end{equation}

\subsubsection{Correction step}
\label{sec:esekf_correction}

In the correction step of the ESEKF, we refine the estimate $ X_k^- $ using the fact that the value $ y_k $ of the measurement at instant $ k $ is known.  Writing the state as $ x = \hat{x} \oplus \delta x $ and linearizing \eqref{eq:measurement_equation} around $ (\hat{x}, 0) $, we obtain
\begin{equation}
    y = g(\hat{x} \oplus \delta x, \nu) \approx g(\hat{x}, 0) \oplus (C(\hat{x}) \delta x + D(\hat{x}) \nu), \label{eq:linearized_correction}
\end{equation}
where $ C(\hat{x}) $ and $ D(\hat{x}) $ are, respectively, $ m \times n $ and $ m \times p $ Jacobian matrices of the function $ g $.

Thus, we can think of $ y $ as being constituted by two components: an expected measurement $ \hat{y} = g(\hat{x}, 0) $ and a perturbation $ \delta y = C(\hat{x}) \delta x + D(\hat{x}) \nu $. Observe that $ \delta y $ can be interpreted as a linear measurement of the error state $ \delta x $ corrupted by additive Gaussian noise. This is exactly the hypothesis of the standard Kalman filter, and we can therefore apply the usual formulas of the correction step to update the estimate $ \delta X_k^- $. Let $ \delta X_k^* $ be this updated estimate, with mean $ \delta \hat{x}_k $ and covariance matrix $ P_k $. Then, we have
\begin{align}
    \delta \hat{x}_k &= K_k (y_k \ominus \hat{y}_k) \\
    P_k &= (I - K_k C_k) P_k^- \label{eq:variance_Pk}
\end{align}
where
$ \hat{y}_k = g(\hat{x}_k^-, 0) $, $ C_k = C(\hat{x}_k^-) $, $ D_k = D(\hat{x}_k^-) $, and $ K_k $ is the Kalman gain, given by
\begin{equation}
    K_k = P_k^- {C_k}^T (C_k P_k^- {C_k}^T + D_k R {D_k}^T)^{-1}.
\end{equation}
The estimate of the state at instant $ k $ is $ X_k = \hat{x}_k^- \oplus \delta X_k^* = \hat{x}_k \oplus \delta X_k $, which has mean
\begin{equation}
    \hat{x}_k = \hat{x}_k^- \oplus \delta \hat{x}_k
\end{equation}
and variance $ P_k $, as in \eqref{eq:variance_Pk}.\footnote{To be rigorous, the variance of $ X_k $ is not exactly that given by \eqref{eq:variance_Pk}, but it should be corrected in an operation known as \textit{error reset}. However, for a small $ \delta \hat{x}_k $, this correction is usually negligible. See \cite{sola2017quaternion} for more details.}

\section{STATE ESTIMATOR EQUATIONS}
\label{sec:algorithm}

We now detail the algorithm described in Section \ref{sec:overview} using the framework developed in Section \ref{sec:esekf}. Consider a humanoid robot instrumented with $ N \ge 3 $ IMUs, each attached to a different link, including, necessarily, both feet. For each link $ i $, we want to estimate its orientation, represented by a rotation matrix $ R_i $, its position $ p_i $, and its linear velocity $ v_i $, all expressed in an inertial reference frame. Each IMU is composed of a gyroscope and an accelerometer, whose measurements $ u_{g_i} $ and $ u_{a_i} $ can be written as
\begin{align}
    u_{g_i} &= \omega_i + b_{g_i} + \eta_{g_i} \label{eq:gyroscope_measurement} \\
    u_{a_i} &= {R_i}^T (\dot{v_i} - g) + b_{a_i} + \eta_{a_i}
\end{align}
where $ \omega_i $ is the body angular velocity in the body frame (i.e., the vector such that $ \dot{R}_i = R_i \ehat{(\omega_i)} $), $ g $ is the gravity vector, $ b_{g_i} $ and $ b_{a_i} $ are unknown biases, and $ \eta_{g_i} $ and $ \eta_{a_i} $ are noise terms. Thus, we take the state of body $ i $ to be $ x_i = (R_i, p_i, v_i, b_{g_i}, b_{a_i}) $, and we denote the corresponding error state by $ \delta x_i = [\delta \theta_i ~ \delta p_i ~ \delta v_i ~ \delta b_{g_i} ~ \delta b_{a_i}]^T $. The full state and error state, $ x $ and $ \delta x $, are the concatenation of the $ x_i $ and $ \delta x_i $ for $ i = 1, ..., N $. Following the notation of Section \ref{sec:esekf}, the state evolves on the 15$N$-dimensional manifold $ \manM_s = (\text{SO(3)} \times \R^{12})^N $, with the $ \oplus $ operator corresponding to the element-wise application of each component's operator.

As explained in Section \ref{sec:overview}, we also have access to two other measurements, which will be used in the correction step: the tilt of a link $ i $, which is defined as $ {R_i}^T e_z \in \Stwo $, $ e_z $ being the unit vector $ [0 ~ 0 ~ 1]^T $ that corresponds to the direction of gravity in the inertial frame \cite{vigne2022movie}, and the relative pose between two links $ i $ and $ j $, which we represent by the ordered pair $ \big( {R_i}^T R_j, {R_i}^T (p_j - p_i) \big) \in \text{SO(3)} \times \R^3 $. We denote
\begin{equation}
    \big( {R_i}^T R_j, {R_i}^T (p_j - p_i) \big) = \text{kin}_{ij}(q)
\end{equation}
where $ \text{kin}_{ij} $ is the forward kinematic function for the relative pose of links $ i $ and $ j $, and $ q $ contains the vector of joint angles and all the rotation matrices that represent the deformations. 

Both these measurements are the result of a non-trivial fusion of the data from the encoders and IMUs. Therefore, to model their uncertainties, we will make several simplifying assumptions.\footnote{A consequence of these assumptions is that we neglect the fact that these measurements are not independent from the gyroscope and accelerometer noise terms that appear in the prediction models.} We suppose that the tilt estimates from MOVIE are corrupted by independent noise terms $ \nu_{t_i} $, and we write their measurements as
\begin{equation}
    y_{t_i} = ({R_i}^T e_z) \oplus \nu_{t_i}. \label{eq:tilt_measurement}
\end{equation}
The uncertainty of the relative poses comes from uncertainties in our estimate of $ q $. Thus, we model their measurements as
\begin{equation}
    y_{r_{ij}} = \text{kin}_{ij}(q \oplus \nu_q)
\end{equation}
where $ \nu_q $ is the vector that contains the noise terms of the joint angles and deformation estimates. Linearizing this equation around $ q $, we obtain the approximation
\begin{equation}
    y_{r_{ij}} \approx \big( {R_i}^T R_j \oplus J^R_{ij} \nu_q, {R_i}^T (p_j - p_i) + J^p_{ij} \nu_q \big) \label{eq:pose_measurement}
\end{equation}
where $ J^R_{ij} $ and $ J^p_{ij} $ are, respectively, the Jacobians of the orientation and position components of $ \text{kin}_{ij} $.

\subsection{Prediction}

As explained in Section \ref{sec:overview}, the prediction model for each link is chosen depending on whether it is classified as a floating link or a contact link at the moment. In either case, the equations that describe the evolution of their state are independent of the other links. This allows us to consider them individually during much of the prediction step. For each link $ i $, we choose a prediction model $ \dot{x}_i = f_i(x_i, \eta_i) $ and integrate it with $ \eta_i = 0 $ to obtain the estimate $ \hat{x}_{i,k}^- $. Then, we compute the differential equation for $ \delta x_i $ and linearize it to obtain the Jacobian matrices $ F_i(\hat{x}_i) $ and $ G_i(\hat{x}_i) $, as in \eqref{eq:error_state_ode}. Finally, substituting $ \hat{x}_i = \hat{x}_{i,k}^- $ and using \eqref{eq:Ak_approx} and \eqref{eq:Qk_approx}, we obtain the discrete-time matrices $ A_{i,k} $ and $ Q_{i,k} $.

The last operation of the prediction step is the computation of the covariance matrix $ P_k^- $. To update the covariance terms between different links correctly, the covariance matrix cannot be computed individually for each link. Thus, we first obtain the full-state matrices $ A_k $ and $ Q_k $ by stacking their link equivalents diagonally, which gives
\begin{align}
    A_k &= \text{diag}(A_{1,k}, ..., A_{N,k}) \\
    Q_k &= \text{diag}(Q_{1,k}, ..., Q_{N,k}),
\end{align}
and then we compute $ P_k^- $ using \eqref{eq:covariance_prediction}.

\subsubsection{Floating link model}

The prediction model used for the floating links is identical to the floating base model from \cite{rotella2014state}. It uses both the gyroscope and accelerometer measurements, and it makes no assumption about the robot's movements. The differential equations for this model are
\begin{align}
    \scriptsize
    \dot{R}_i   &= R_i \ehat{(u_{g_i} - b_{g_i} - \eta_{g_i})} \label{eq:prediction_floating_R}\\
    \dot{p}_i   &= v_i \\
    \dot{v}_i   &= R_i (u_{a_i} - b_{a_i} - \eta_{a_i}) + g \\
    \dot{b}_{g_i} &= \eta_{b_{g_i}} \label{eq:prediction_floating_bg} \\
    \dot{b}_{a_i} &= \eta_{b_{a_i}} \label{eq:prediction_floating_ba}
\end{align}
where the bias terms are modelled as Brownian motions with derivatives $ \eta_{b_{g_i}} $ and $ \eta_{b_{a_i}} $.

Under a zero-order hold assumption for the sensors, we can compute the estimate $ \hat{x}_{i,k}^- $ as
\begin{align}
    \hat{R}_{i,k}^- &= \hat{R}_{i,k-1} \Exp{[(u_{g_i, k-1} - \hat{b}_{g_i, k-1}) \Delta t]} \\
    \hat{p}_{i,k}^- &= \hat{p}_{i,k-1} + \hat{v}_{i,k-1} \Delta t \notag \\ &\qquad + [\hat{R}_{i,k-1}(u_{a_i, k-1} - \hat{b}_{a_i, k-1}) + g] \frac{\Delta t^2}{2} \\
    \hat{v}_{i,k}^- &= [\hat{R}_{i,k-1}(u_{a_i, k} - \hat{b}_{a_i, k-1}) + g] \Delta t \\
    \hat{b}_{g_i, k}^- &= \hat{b}_{g_i, k-1} \\
    \hat{b}_{a_i, k}^- &= \hat{b}_{a_i, k-1}.
\end{align}
Then, we compute the continuous-time linearized differential equations for the error state, which are
\begin{align}
    \delta \dot{\theta}_i &= -\ehat{(u_{g_i} - \hat{b}_{g_i})} \delta \theta_i - \delta b_{g_i} - \eta_{g_i} \\
    \delta \dot{p}_i &= \delta v_i \\
    \delta \dot{v}_i &= -\hat{R}_i \ehat{(u_{a_i} - \hat{b}_{a_i})} \delta \theta_i - \hat{R}_i \delta b_{a_i} - \hat{R}_i \eta_{a_i} \\
    \delta \dot{b}_{g_i} &= \eta_{b_{g_i}} \\
    \delta \dot{b}_{a_i} &= \eta_{b_{a_i}}.
\end{align}
From these equations, we extract the Jacobian matrices
\begin{gather}
F_i(\hat{x}_i) =
\begin{pmatrix}
    -\ehat{(u_{g_i} -\hat{b}_{g_i})}            & 0 & 0 & -I &  0         \\
     0                                          & 0 & I &  0 &  0         \\
    -\hat{R}_i \ehat{(u_{a_i} - \hat{b}_{a_i})} & 0 & 0 &  0 & -\hat{R}_i \\
     0                                          & 0 & 0 &  0 &  0         \\
     0                                          & 0 & 0 &  0 &  0
\end{pmatrix}\\
G_i(\hat{x}_i) =
\begin{pmatrix}
    -I &  0         & 0 & 0 \\
     0 &  0         & 0 & 0 \\
     0 & -\hat{R}_i & 0 & 0 \\
     0 &  0         & I & 0 \\
     0 &  0         & 0 & I
\end{pmatrix}
\end{gather}
for which the noise vector is $ \eta_i = [\eta_{g_i} ~ \eta_{a_i} ~ \eta_{b_{g_i}} ~ \eta_{b_{a_i}}]^T $.

\subsubsection{Contact link model}

Instead of relying on the integration of the accelerometer measurements to compute the velocity of the contact links, we consider them to be instantly rotating around their center of pressure. As such, we write
\begin{equation}
    v_i = {}^\mathcal{W}\omega_i \times {}^\mathcal{W}r_i + v_{COP_i} \label{eq:contact_velocity}
\end{equation}
where $ v_{COP_i} $ is the velocity of the center of pressure, $ {}^\mathcal{W} r_i $ is the position vector of the link's origin with respect to the center of pressure, and the superscript $ {}^\mathcal{W} $ indicates that the variables are expressed in the inertial reference frame. We assume the velocity of the center of pressure is approximately zero, and we model it as a zero-mean Gaussian random variable $ \eta_{s_i} $ to account for a small slippage. Observe that the angular velocity measured by the gyroscope is expressed in the link frame, and it is therefore different from $ {}^\mathcal{W} \omega_i $; similarly, we can compute $ r_i $ in the link frame, but we depend on the orientation $ R_i $ to obtain $ {}^\mathcal{W} r_i $. Given these considerations, we rewrite \eqref{eq:contact_velocity} as
\begin{equation}
    v_i = (R_i \omega_i) \times (R_i r_i) + \eta_{s_i}.
\end{equation}

The equations for the link orientation and the biases are the same as for the floating links. The complete prediction model for the contact links is therefore composed of equations \eqref{eq:prediction_floating_R}, \eqref{eq:prediction_floating_bg}, \eqref{eq:prediction_floating_ba}, and
\begin{equation}
    \dot{p_i} = v_i = R_i [(u_{g_i} - b_{g_i} - \eta_{g_i}) \times r_i] + \eta_{s_i}
\end{equation}
where we used the fact that multiplication by a rotation matrix is distributive with respect to the cross product and substituted the expression for $ \omega_i $ of \eqref{eq:gyroscope_measurement}.

An important difference with respect to the floating link model is there is no equation for $ \dot{v}_i $. Nevertheless, we would like to continue estimating the covariance terms that involve $ v_i $, because the contact link will become a floating link once it leaves the floor. We will therefore proceed with a reduced state $ x_i = (R_i, p_i, b_{g_i}, b_{a_i}) $ up to the computation of the matrices $ A_{i,k} $ and $ Q_{i,k} $ and then simply add to them the rows and columns corresponding to $ v_i $, whose elements can be obtained directly from \eqref{eq:contact_error_velocity}. 

Following the same procedure as before, the estimate $ \hat{p}_{i,k}^- $ is computed as
\begin{equation}
    \hat{p}_{i,k}^- = \hat{p}_{i,k-1} + \hat{R}_{i,k-1} [(u_{g_i,k-1} - b_{g_i,k-1}) \times r_i] \Delta t
\end{equation}
and the linearized model for the evolution of its error is
\begin{multline}
    \delta \dot{p}_i = \delta v_i =  -\hat{R}_i \ehat{[(u_{g_i} - \hat{b}_{g_i}) \times r_i]} \delta \theta_i \\+ \hat{R}_i \ehat{(r_i)} \delta b_{g_i} + \hat{R}_i \ehat{(r_i)} \eta_{g_i} + \eta_{s_i} \label{eq:contact_error_velocity}.
\end{multline}
The Jacobian matrices for the contact link model are then
\begin{equation}
\resizebox{0.89\linewidth}{!}{$%
F_i(\hat{x}_i) =
\begin{pmatrix}
    -\ehat{(u_{g_i} - \hat{b}_{g_i})}                        & 0 & -I                     & 0 \\
    -\hat{R}_i \ehat{[(u_{g_i} - \hat{b}_{g_i}) \times r_i]} & 0 & \hat{R}_i \ehat{(r_i)} & 0 \\
     0                                                       & 0 &  0                     & 0 \\
     0                                                       & 0 &  0                     & 0
\end{pmatrix}$}
\end{equation}

\begin{equation}
G_i(\hat{x}_i) =
\begin{pmatrix}
    -I                      & 0 & 0 & 0 \\
     \hat{R}_i \ehat{(r_i)} & 0 & 0 & I \\
     0                      & I & 0 & 0 \\
     0                      & 0 & I & 0
\end{pmatrix}
\end{equation}
for which the noise vector is $ \eta_i = [\eta_{g_i} ~ \eta_{b_{g_i}} ~ \eta_{b_{a_i}} \eta_{s_i}]^T $.

\subsection{Correction}

After the state estimates of all bodies have been updated in the prediction step, we correct them using the tilt and relative pose measurement values. Because the noise terms of these two types of measurements were modelled as independent from each other, we can consider them separately and perform the updates to the state estimates sequentially.

\subsubsection{Tilt measurements}

 We follow the procedure detailed in Section \ref{sec:esekf_correction} for a measurement defined on the manifold $ \manM_m = \Stwo $. For each contact link $ i $, we compute the Jacobian matrices $ C_{t_i}(\hat{x}) $ and $ D_{t_i}(\hat{x}) $ corresponding to the linearization of the measurement $ y_{t_i} $ as in \eqref{eq:linearized_correction}, and we apply all the correction equations of the ESEKF. 

The linearized measurement equation is
\begin{equation}
    \delta y_{t_i} = B(\hat{R}_i)^T [\ehat{({\hat{R}_i}^T e_z)}]^2 (\delta \theta_i - B(\hat{R}_i) \nu_{t_i})
\end{equation}
where $ B(\hat{R}_i) $ is a $ 3 \times 2 $ matrix whose columns are the first two rows of $ \hat{R}_i $. It fulfills the role of matrix $ B(u) $ of \eqref{eq:oplus_s2}, as its columns span the subspace orthogonal to the tilt $ {\hat{R}_i}^T e_z $.
This gives us the Jacobians
\begin{equation}
    C_{t_i}(\hat{x}) =
    \begin{pmatrix}
        \ldots & B(\hat{R}_i)^T [\ehat{({\hat{R}_i}^T e_z)}]^2 & \ldots
    \end{pmatrix}
\end{equation}
\begin{equation}
    D_{t_i}(\hat{x}) = - B(\hat{R}_i)^T [\ehat{({\hat{R}_i}^T e_z)}]^2 B(\hat{R}_i)
\end{equation}
where the only non-zero block of $ C_{t_i}(\hat{x}) $ occupies the columns corresponding to $ \delta \theta_i $.

\subsubsection{Relative pose measurements}

Because the uncertainties of all relative pose measurements originate from the same noise term $ \nu_q $, we will consider them simultaneously when correcting the state estimate. Hence, we follow the steps of Section \ref{sec:esekf_correction} for $ \manM_m = (\text{SO(3)} \times \R^3)^d $, $ d $ being the number of pairs of floating and contact links.

For each contact link $ i $ and floating link $ j $ pair, we compute the Jacobians $ C_{r_{ij}}(\hat{x}) $ and $ D_{r_{ij}}(\hat{x}) $ from the linearization of $ y_{r_{ij}} $. The linearized measurement equation is
\begin{equation}
\resizebox{0.89\linewidth}{!}{$%
    \delta y_{r_{ij}} = 
    \begin{pmatrix}
        -{\hat{R}_j}^T \hat{R}_i \delta \theta_i + \delta \theta_j + J^R_{ij} \nu_q \\
        \ehat{[{\hat{R}_i}^T(\hat{p}_j - \hat{p}_i)]} \delta \theta_i - {\hat{R}_i}^T \delta p_i + {\hat{R}_i}^T \delta p_j + J^p_{ij} \nu_q
    \end{pmatrix}$}
\end{equation}
from which we obtain
\begin{equation}
\resizebox{0.89\linewidth}{!}{$%
    C_{r_{ij}}(\hat{x}) =
    \begin{pmatrix}
        \ldots & -{\hat{R}_j}^T \hat{R}_i                      &  0             & \ldots & I & 0             & \ldots \\
        \ldots & \ehat{[{\hat{R}_i}^T(\hat{p}_j - \hat{p}_i)]} & -{\hat{R}_i}^T & \ldots & 0 & {\hat{R}_i}^T & \ldots
    \end{pmatrix}$}
\end{equation}
\begin{equation}
    D_{r_{ij}}(\hat{x}) =
    \begin{pmatrix}
        J^R_{ij} \\
        J^p_{ij}
    \end{pmatrix}.
\end{equation}
The columns occupied by the first block of matrix $ C_{r_{ij}}(\hat{x}) $ correspond to $ \delta \theta_i $ and $ \delta p_i $, and those of the second to $ \delta \theta_j $ and $ \delta p_j $. Then, it suffices to stack together the rows of the matrices $ C_{r_{ij}}(\hat{x}) $ and those of the matrices $ D_{r_{ij}}(\hat{x}) $ to obtain the Jacobians $ C_r(\hat{x}) $ and $ D_r(\hat{x}) $ of the full set of relative pose measurements before proceeding to the computation of the Kalman gain.

\section{EXPERIMENTAL RESULTS}
\label{sec:results}

\begin{table*}[t]
    \centering
    \caption{Relative pose error (RPE), absolute trajectory error (ATE) and average vertical drift per step (AVDS) of the exoskeletons' pelvis trajectory for both experiments. The ATE is the RMS error of the whole trajectory, the RPE is the median of the relative errors computed with a sliding window of 0.5 s, and the AVDS is the total error in the Z direction divided by the number of steps taken.}
    \small
    \begin{tabularx}{0.9\textwidth}{ c *{6}{Y} }
    \toprule
                        & \multicolumn{3}{c}{\textbf{Exp. 1 - Straight trajectory}}  & \multicolumn{3}{c}{\textbf{Exp. 2 - Circular trajectory}}  \\
    \cmidrule(lr){2-4} \cmidrule(l){5-7}
                        & \textbf{RPE (cm)} & \textbf{ATE (cm)} & \textbf{AVDS (mm)} & \textbf{RPE (cm)} & \textbf{ATE (cm)} & \textbf{AVDS (mm)} \\
    \midrule
    \textbf{1-IMU}      & 1.4               & 10.2              & 3.0                & 3.6               & 19.3              & 3.4                \\
    \textbf{1-IMU-EKM}  & 1.0               & 11.5              & 2.2                & 3.0               & 21.4              & 3.8                \\
    \textbf{5-IMU-EKM}  & \textbf{0.6}      & \textbf{5.9}      & \textbf{0.3}       & \textbf{2.5}      & \textbf{14.9}     & \textbf{0.3}       \\
    \bottomrule
    \end{tabularx}
    \label{tab:experimental_metrics}
\end{table*}

The proposed multi-IMU state estimator was tested on data collected from Atalante, a self-balanced leg exoskeleton that allows patients with motor disabilities to walk without crutches. Atalante is instrumented with an encoder on each of its 12 joints, five low-cost IMUs (LSM6DSO)  --- one on the pelvis, one on each tibia and one on each foot, as shown in Fig. \ref{fig:atalante_heel_toe} ---, and four 1D force sensors under each foot. The structure of the exoskeleton deforms   at the hip and ankle joints due to its own weight and that of the patient. Using the available IMUs, MOVIE \cite{vigne2022movie} can estimate these four deformations. All sensors are sampled at 1 kHz, and the filter is run at this same frequency. The estimation runs in real-time on a Kontron 3.5$^{\prime\prime}$-SBC-TGL single-board computer.

Two experiments were conducted, in which we compare our method with two variations of the single-IMU filter \cite{rotella2014state}. The multi-IMU filter, denoted as 5-IMU-EKM, estimates the pose of all five IMUs and uses an extended kinematic model with the four aforementioned deformations. For this filter, a foot is considered in contact with the ground as soon as it touches it, i.e., if the reading of at least one of its force sensors exceeds a threshold. The force sensors are also used to compute the position of the center of pressure. The single-IMU filters, on the other hand, only estimate the pose of the pelvis IMU. The two variations, denoted as 1-IMU and 1-IMU-EKM, differ in the kinematic model employed: 1-IMU uses the standard rigid kinematic model, while 1-IMU-EKM uses the same extended kinematic model as 5-IMU-EKM with four deformations estimated by MOVIE. To maintain consistency with the constant-pose assumption of the stance feet, the single-IMU filters only consider a foot to be in contact with the ground if the readings of two diagonally opposed force sensors exceed a threshold, which indicates that the contact is flat. The performance of the filters is evaluated based on the absolute trajectory error (ATE) and relative pose error (RPE) \cite{sturm2012benchmark} of the pelvis trajectory, with ground truth obtained from an OptiTrack motion capture system. The values obtained for these metrics in the two experiments are given in Table \ref{tab:experimental_metrics}, in which we also report the average drift per step in the vertical direction (hereafter denoted as AVDS).

\begin{figure*}
    \begin{subfigure}[c]{0.49\textwidth}
        \centering
        \includegraphics[width=\linewidth]{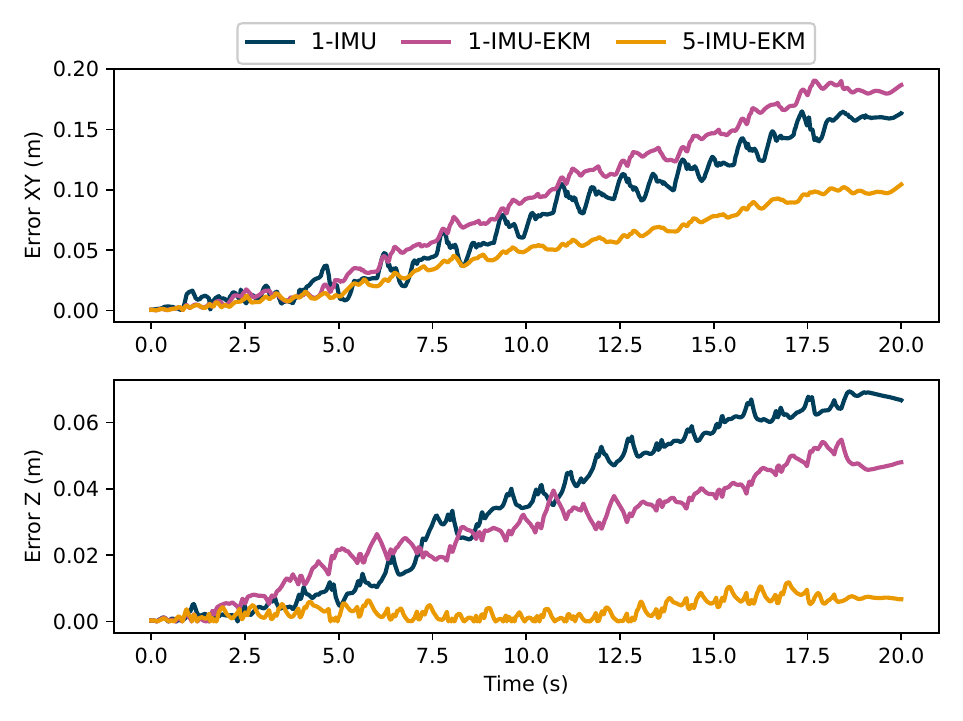}
        \caption{Experiment 1 - Straight trajectory}
        \label{fig:position_errors_early_gait}
    \end{subfigure}
    \hspace{0cm}
    \begin{subfigure}[c]{0.49\textwidth}
        \centering
        \includegraphics[width=\linewidth]{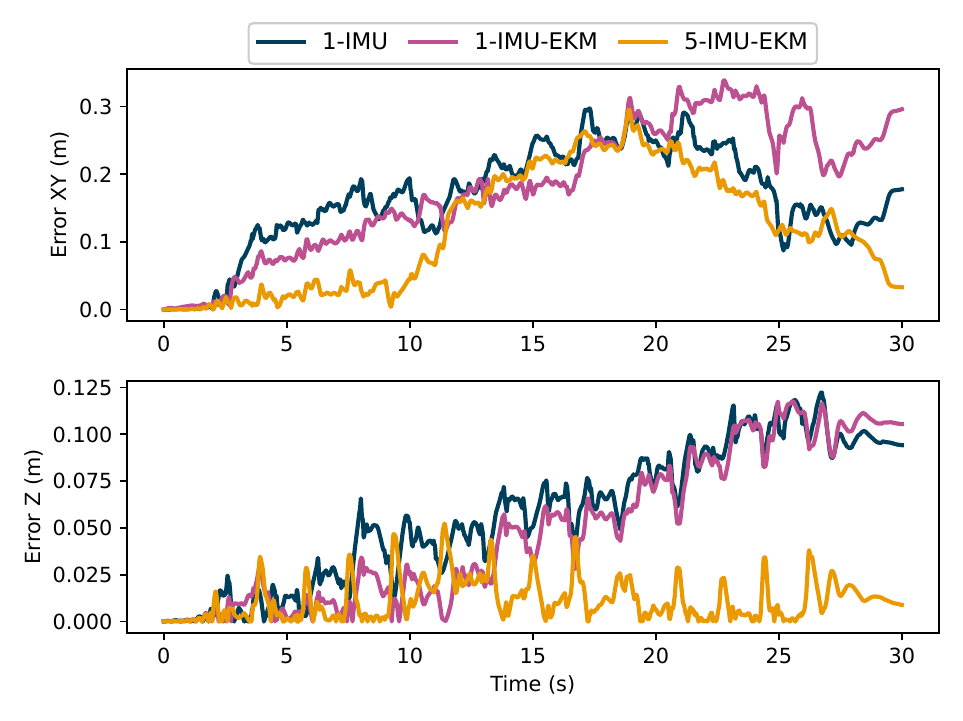}
        \caption{Experiment 2 - Circular trajectory}
        \label{fig:position_errors_real_gait}
    \end{subfigure}
    \caption{Comparison of the position errors, i.e., the Euclidan distance between the trajectory estimate and the ground truth, in the XY plane and in the Z direction for the two performed experiments.}
    \label{fig:my_label}
\end{figure*}

In the first experiment, the exoskeleton walks forward in a straight trajectory at a speed of 0.15 m/s for 3 m, which is the longest distance that can be covered with our motion capture setup. For this gait pattern, the foot touches the floor perfectly flat, but the heel leaves the floor slightly before the toes. The position errors from the three filters are shown in Fig. \ref{fig:position_errors_early_gait}. We observe that the multi-IMU filter outperforms the two single-IMU filters according to all metrics. The difference in their accuracy is especially remarkable in the Z direction, in which the AVDS of 5-IMU-EKM is one order of magnitude lower than that of the others. The orientation errors were smaller than $ 0.5 \degree $ for all three filters. Using the extended kinematic model improved the performance of the single-IMU filter with respect to RPE and AVDS, but not with respect to ATE. This indicates that the accuracy gained with the proposed multi-IMU approach cannot be explained exclusively by a different kinematic model and must therefore be due to the filter formulation.

In the second experiment, a more dynamic gait pattern with significant heel-toe motion is considered. The exoskeleton walks forward at a speed of 0.4 m/s and is pushed to make one and a half turns of an approximately circular trajectory, for a total walking distance of 11.5 m. The position errors for this experiment are presented in Fig. \ref{fig:position_errors_real_gait}. The three metrics show that 5-IMU-EKM also performs better than the other filters in this case, especially regarding the accuracy in the Z direction. The multi-IMU filter also presented less drift in the yaw angle: the final error was approximately equal to $ 2 \degree $, in contrast to more than $ 6 \degree $ for the single-IMU filters.  Once again, these differences cannot be simply attributed to the extended kinematic model. We draw attention to the fact that even though this more dynamic motion induces errors in all three filters, the AVDS of 5-IMU-EKM is the same as that in the first experiment. This suggests that the proposed filter formulation correctly captures the motion of the feet, even in the presence of heel-toe motion.

\begin{figure*}
    \centering
    \begin{subfigure}{0.22\linewidth}
        \centering
        \includegraphics[height=120pt]{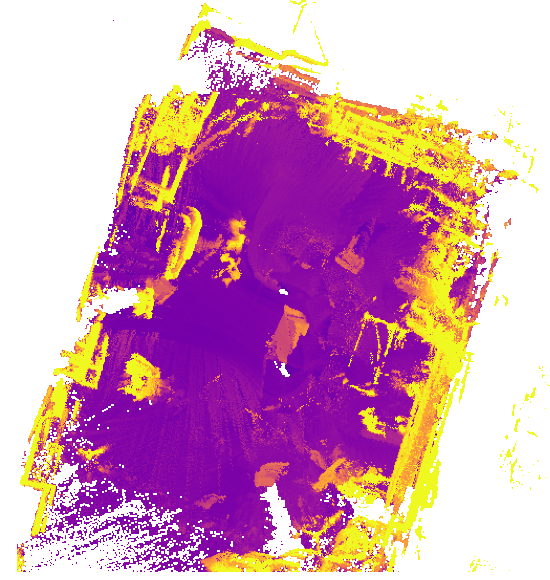}
        \caption{1-IMU}
    \end{subfigure}
    \hfill
    \begin{subfigure}{0.22\linewidth}
        \centering
        \includegraphics[height=120pt]{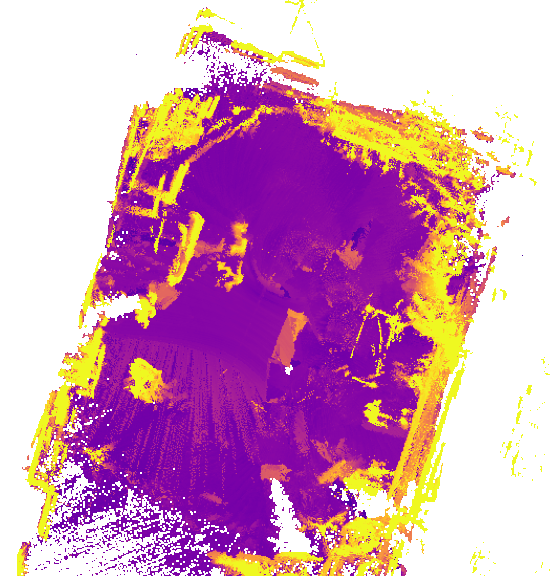}
        \caption{1-IMU-EKM}
    \end{subfigure}
    \hfill
    \begin{subfigure}{0.22\linewidth}
        \centering
        \includegraphics[height=120pt]{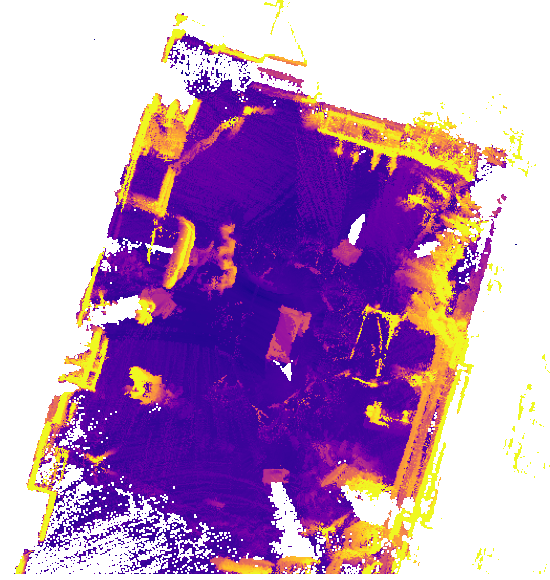}
        \caption{5-IMU-EKM}
    \end{subfigure}
    \hfill
    \begin{subfigure}{0.22\linewidth}
        \centering
        \includegraphics[height=115pt]{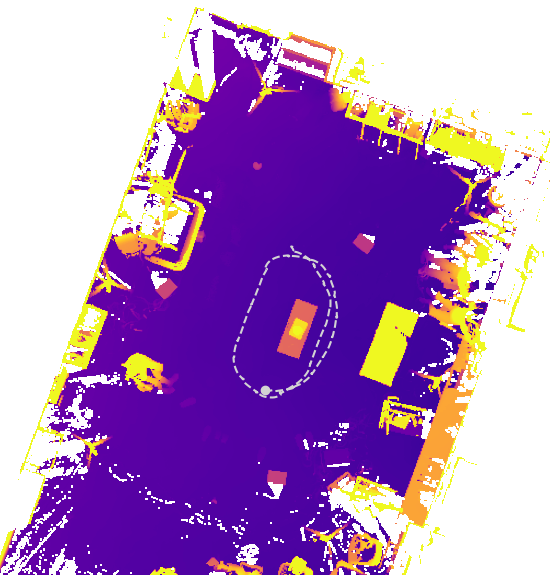}
        \caption{Ground truth}
    \end{subfigure}
    \hfill
    \begin{subfigure}{0.08\linewidth}
        \centering
        \includegraphics[height=120pt]{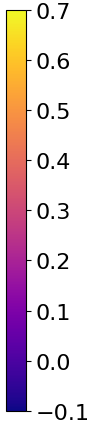}
    \end{subfigure}
    \caption{Elevation maps constructed from the filter trajectories and ground truth obtained using a Faro laser scanner. The trajectory of the exoskeleton is depicted as a dashed line in the ground truth map. The heights are clipped between -0.1 m and 0.7 m for visualization purposes. Some objects present in the ground truth map do not appear in the other elevation maps because they were outside the LiDAR's field of view.}
    \label{fig:elevation_maps}
\end{figure*}

To illustrate the potential of the multi-IMU state estimator, a Velarray M1600 solid-state LiDAR was attached to the exoskeleton's right knee for this second experiment. Based on the trajectory estimate of each filter, an elevation map of the surroundings of the exoskeleton was constructed using a simplified implementation of the algorithm \cite{fankhauser2018probabilistic}, which does not perform scan matching. The resulting maps are shown in Fig. \ref{fig:elevation_maps}, where they are compared to the terrain ground truth obtained using a Faro Focus3D-X130 laser scanner.  The difference in the accuracy of the trajectory estimates from the filters is also clearly visible in the elevation maps. The drift in the Z direction of the single-IMU filters creates height differences with respect to the ground truth, indicated by the different color of the floor between the maps, and the errors in yaw and in the XY plane cause the edges of the map to be misaligned. The elevation map from the multi-IMU filter, conversely, resembles the ground truth more closely, with respect to both its height values and the alignment of its edges. 

\section{Conclusion}

In this paper, we presented a novel EKF-based state estimator for humanoid robots that can provide low-drift trajectory estimates even when walking with strong heel-toe motion. The filter leverages data from multiple IMUs to model the motion of different links, including the robot's feet, which allows us to drop the common assumption that their pose remains constant while in contact with the ground. This sensor configuration is also used to estimate the deformations in the mechanical structure, improving the kinematic model used by the filter. The proposed multi-IMU estimator was tested on the exoskeleton Atalante with both slow and dynamic gaits, outperforming other single-IMU estimators. When using the obtained trajectory estimates to construct elevation maps of the surroundings of the exoskeleton, only the map based on the multi-IMU estimator was accurate enough for potential use in motion planning and navigation tasks.

\addtolength{\textheight}{-1.3cm}   




\bibliographystyle{IEEEtran}
\bibliography{references.bib}

\end{document}